# Re-assembling the past: The RePAIR dataset and benchmark for real world 2D and 3D puzzle solving

**Theodore Tsesmelis**[1]⋆  **Luca Palmieri**[2]⋆  **Marina Khoroshiltseva**[2]  **Adeela Islam**[1]
**Gur Elkin**[3]  **Ofir Itzhak Shahar**[3]  **Gianluca Scarpellini**[1]  **Stefano Fiorini**[1]  **Yaniv Ohayon**[3]
**Nadav Alali**[3]  **Sinem Aslan**[2,7]  **Pietro Morerio**[1]  **Sebastiano Vascon**[2]  **Elena Gravina**[4]
**Maria Cristina Napolitano**[4]  **Giuseppe Scarpati**[4]  **Gabriel Zuchtriegel**[4]  **Alexandra Spühler**[5]
**Michel E. Fuchs**[5]  **Stuart James**[1,6]  **Ohad Ben-Shahar**[3]  **Marcello Pelillo**[2]  **Alessio Del Bue**[1]
[1]Fondazione Istituto Italiano di Tecnologia  [2]Ca' Foscari University of Venice
[3]Ben-Gurion University of the Negev  [4]Parco Archeologico di Pompei
[5]University of Lausanne  [6]Durham University  [7]University of Milan

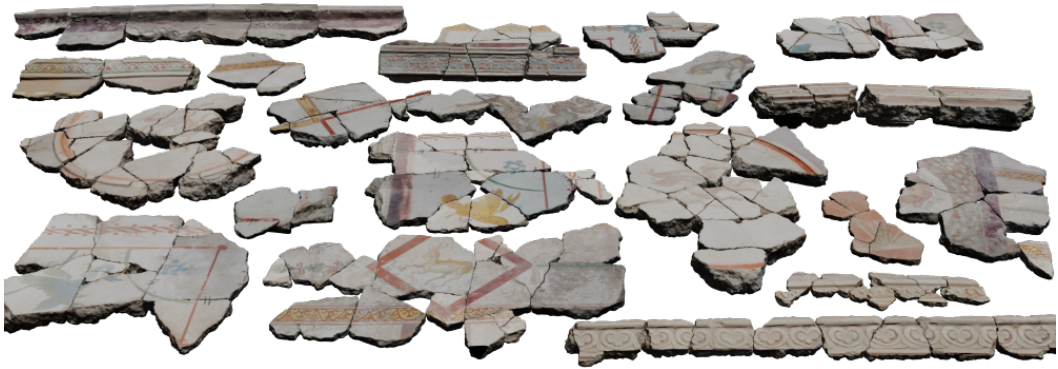

Figure 1: A subset of the RePAIR dataset, containing over one thousand real-life object fragments. The dataset is unique in its doing, with an international team of archaeologists providing the puzzle ground truth through years of fieldwork in the Pompeii archaeological park. This dataset stands as a realistic computational challenge for both methods on 2D and 3D puzzle solving, as modern baselines struggle to achieve competitive results on the RePAIR dataset.

## Abstract

This paper proposes the RePAIR dataset that represents a challenging benchmark to test modern computational and data driven methods for puzzle-solving and reassembly tasks. Our dataset has unique properties that are uncommon to current benchmarks for 2D and 3D puzzle solving. The fragments and fractures are realistic, caused by a collapse of a fresco during a World War II bombing at the Pompeii archaeological park. The fragments are also eroded and have missing pieces with irregular shapes and different dimensions, challenging further the reassembly algorithms. The dataset is multi-modal providing high resolution images with characteristic pictorial elements, detailed 3D scans of the fragments and meta-data annotated by the archaeologists. Ground truth has been generated through several years of unceasing fieldwork, including the excavation and cleaning of each fragment, followed by manual puzzle solving by archaeologists of a subset of approx. 1,000 pieces among the 16,000 available. After digitizing all the fragments in 3D, a benchmark was prepared to challenge current reassembly and puzzle-solving methods that often solve more simplistic synthetic scenarios. The tested baselines show that there clearly exists a gap to fill in solving this computationally complex problem.

---

⋆: equal contribution

# 1 Introduction

Archaeologists and historians face the very same problem of reconstructing the past. Unlike historians, though, for archaeologists, the reconstruction process has almost invariably a physical dimension. Indeed, after a long and painstaking work involving site surveys and excavations, archaeologists are often confronted with the challenging problem of reassembling countless fragments of different size, shape and appearance to recreate ancient artifacts or artworks. Depending on the size of the artifact, the reassembly work might last months, years, or even decades, and in the presence of a large number of pieces, it simply becomes hopeless, notwithstanding the skillfulness of the operators. It is, therefore, quite common for many museums and archaeological parks to set up large storerooms to stow hundreds or thousands of such fragments, just waiting for someone (or something) to reassemble them for the benefit of scholars, laypersons, and our collective cultural heritage alike.

We, therefore, frame the RePAIR dataset to address the computational challenge of assembling frescoes from their constituent fragments. Together with the UNESCO World Heritage site of Pompeii we created a unique dataset of an iconic case study of large-scale fragmented painted wall fresco. We meticulously 3D reconstruct the fragments using State-of-The-Art (SoTA) Computer Vision techniques, providing both sub-millimeter accuracy for geometry and high-resolution texture details. Then, with the support of an international team of archaeologists, we provide a ground-truth solution to the 2D and 3D puzzle-solving problems that are at the basis of our benchmark. Unlike prior datasets, the RePAIR is a real-world dataset of greater scale and complexity that offers new challenges to the machine learning and computer vision communities.

The RePAIR dataset provides real complex geometries, large variations in fracture volumes due to physical causes, and differing numbers of fractured and missing pieces per shape as shown in Fig. 1. The 1070 digitized pieces have been manually assembled into 117 coherent groups by the team of archaeologists and used to provide a set of equivalent puzzles to solve, both in 2D and 3D. In our dataset, we do not only provide the puzzle solutions in terms of the translation and rotation of each fragment but also augment each piece with archaeologists-provided metadata. For instance, the distinct 3D structures on the back sides of fragments, reflecting their historical use, offer alignment cues. For this reason, we annotate each object with variations in material features—such as material, weight, dimensions, and size—serve as additional indicators of successful matches.

We analyze RePAIR by introducing standardized metrics for puzzle-solving and reassembly and we benchmark several state-of-the-art models under various settings. Extensive experiments against the RePAIR dataset reveal that fractured shape reassembly is still an open problem, especially when methods are confronted with a realistic task as represented by the RePAIR dataset.

**Summary of contributions:** *i)* We reconstruct high-quality 3D fresco fragments and include corresponding 2D high-quality rendered image of the painted surface of each fresco fragment, offering a unique link between a 3D and a 2D dataset. *ii)* We introduce a large-scale dataset of *irregular* fractured objects for the puzzle solving task both in 2D and 3D. *iii)* We provide a geometric analysis of the digitized fragments and include this information as metadata files. *iv)* We propose two approaches for the *irregular* 2D puzzle-solving task, for which only one previous method was available. *v)* We benchmark SoTA methods for puzzle solving in 2D and 3D on our dataset, with open source code to ensure reproducibility and facilitate future research. *vi)* We will publicly release this dataset, including images and 3D models, to facilitate comparative studies in fragment registration and reconstruction methods, as well as related research, such as 3D reassembly and restoration.

# 2 Related Datasets and Benchmarks

Here, we introduce the background for 2D and 3D datasets (Sec. 2.1) as well as current SoTA approaches for both solving various forms of puzzles (Sec. 2.2).

## 2.1 Datasets for 2D and 3D puzzle-solving

**Datasets for 2D puzzle solving.** Datasets for 2D puzzle solving exist as either real scanned fragments or synthesized from images. Synthetic approaches can generate a wide range of puzzle types and geometries, starting from square pieces [16, 64, 78], moving to polygonal fragments [33, 34] and expanding to more complex and naturalistic-looking shapes. In the Crossing Cuts [34] and DAFNE datasets [22], polygonal parameterized procedures were used to break images into pieces, while

Table 1: Publicly available 2D puzzle solving datasets and their properties.

| Dataset | No. Groups | No. Breaks | Breakdown Type | Eroded? | Real/Synthetic | Image Origin |
|---|---|---|---|---|---|---|
| Theran [9] | N/A | 44-129 | Unrestricted | ✓ | Both | Frescoes |
| DAFNE [22] | 62 | 100-2700 | Unrestricted | ✓ | Synthetic | Frescoes |
| Derech et al. [19] | 31 | 10-40 | Unrestricted | ✓ | Synthetic | Frescoes |
| Crossing Cuts [34] | 7200 | 10-3907 | Crossing Cuts, Square, Perturbed | ✓ | Synthetic | Natural Images |
| Dead Sea Scrolls [46] | 5 | N\A | Unrestricted | ✓ | Real | Scrolls |
| Church of Antiphonitis [2] | 2 | 32-36 | Unrestricted | ✓ | Real | Frescoes |
| Small Temple of Petra [52] | 1 | 78 | Unrestricted | ✓ | Real | Frescoes |
| Ovetari Chapel [26] | 1 | 80735 | Unrestricted | ✓ | Real | Frescoes |
| Sholomon et al. [72] | 20 | 5015-22834 | Square | ✗ | Synthetic | Natural Images |
| Cho et al. [16] | 20 | 432 | Square | ✗ | Synthetic | Natural Images |
| Pomeranz et al. [64] | 26 | 805-3300 | Square | ✗ | Synthetic | Natural Images |
| Dunhuang Scrolls [1] | N/A | 10-60 | Unrestricted | ✓ | Real | Scrolls |
| RePAIR | 101* | 2-44 | Unrestricted | ✓ | Real | Frescoes |

* Some fragments are decorated with *Stucco* and thus lack a planar surface to be transformed to 2D. Hence, fewer fragments in 2D (See Sec. 3.3).

imitating the wear and tear of real archaeological puzzles through a simulated erosion function. This approach enabled the creators of these datasets to generate puzzles with a varying number of pieces and an exact ground truth – a crucial feature for many machine learning implementations and a basic tool for evaluating solvers. A similar yet different method was used to generate training data for JigsawNet [44], where puzzles were created by slicing images using irregular curves, and the task was the reassembly of shredded documents. A more pattern-based synthesis approach was used by Derech et al. [20], where the authors employed natural patterns of dry mud to generate puzzles from existing images of historic wall paintings. It was argued that the shape and gaps between mud pieces were analogous to the fragmentation and erosion of ancient artifacts.

Datasets of scanned fragments provide more realistic data and more genuine puzzle solving challenges but the scanning process is a labor-intensive task, often with limited examples, and almost exclusively in 3D. Datasets in 2D are typically a byproduct of the 3D scanning, perhaps best exemplified in the Theran dataset [9] of Bronze Age wall paintings from Thera (modern-day Santorini). The RePAIR dataset belongs to this type of 2D collections, where it is the most extensive with over 100 different puzzles. Table 1 lists many synthetic and real (scanned) 2D datasets available in the literature.

**Datasets for 3D puzzle solving** The available real-world 3D datasets are confined to scanning multiple fragments corresponding to a limited selection of initially intact objects [38, 79, 87] or to a limited number of breakdown patterns [43]. Previous datasets in [38] and [79] feature only 7 objects, while [87, 57] only 15 and 18 with total broken pieces for each to be counted in 101, 69, 123 and 103 pieces respectively. [43] tried to fill this gap by publishing a relatively larger scale of objects, *i.e.* 150, but their breakdown pattern is limited to two fractured pieces per object. As fracture acquisition occurs post-damage for historical objects without known counterparts, the datasets lack comprehensive knowledge of the complete proxies. This is also the case for the Theran frescoes dataset [9], which reports a total amount of 412 frescoes without being able to report the originating complete structure. The RePAIR dataset is comparable to these prior datasets while extending the number of total amount of pieces to more than double of the maximum reported amount of pieces (see Tab. 2). Acknowledging the necessity for extensive datasets to facilitate learning-driven repair processes, a handful of datasets incorporate synthetic or scanned models exposed to synthetic fractures via geometric methodologies like subtracting primitives [13] or employing physics-based fracture models [70]. In all instances of these datasets though, mid-scale detail is depicted as cutouts using analytical primitives, a representation method that lacks generalization to real-world fracture damage.

## 2.2 Puzzle solving methods

**Puzzle solving in 2D.** The challenge of solving 2D jigsaw puzzles, or reconstructing a coherent whole from an unordered set of fragments, has fascinated humanity for generations. It was first introduced as a *computational* task in 1964 by Freeman and Garder [27]. Since the problem has been proven to be NP-complete [18], research focuses on heuristics, specialized techniques, and computational strategies. Approaches do not guarantee an optimal solution but are frequently successful.

However, most of these puzzle-solving methods relied on heavy constraints regarding shape and other properties of puzzle pieces, or prior knowledge of the ground truth pictorial solution, *a.k.a.*

Table 2: Analysis of the current 3D reassembly and puzzle datasets in the literature.

| Datasets | No. Pieces | No. Breaks (Avg.) | No. Groups | Type | Breakdown Type | Metadata | Texture | Availability | Data Type |
|---|---|---|---|---|---|---|---|---|---|
| Tuwien [38] | 101 | 6-30 (14.42) | 7 | Artifacts | Real | ✗ | ✗ | ✓ | 3D |
| Theran [9] | 412 (283/129)* | - | - | Frescoes | Real | ✗ | ✗ | ✗ | 2D/3D |
| Presious [79] | 69 | 3-30 (9.85) | 7 | Artifacts | Real | ✗ | ✗ | ✓ | 3D |
| Breaking Bad [70] | 8442044 | 2-100 (8.06) | 10474 | Groups | Synthetic | ✗ | ✗ | ✓ | 3D |
| Fantastic Breaks [43] | 300 | 2 (2) | 150 | Groups | Real | ✗ | ✗ | ✓ | 3D |
| FIRES [87] | 123 | 7-18 (8.2) | 15 | Pottery | Real | ✗ | ✗ | ✓ | 3D |
| Chinaware Fragments [57] | 103 | N/A | 18 | Pottery | Real | ✗ | ✗ | ✗ | 3D |
| **RePAIR** | **1070 (951/119)\*** | **2-44 (11.88)** | **117\*\*** | **Frescoes** | **Real** | ✓ | ✓ | ✓ | **2D/3D** |

* Naturally/Artificially made     ** No. Puzzles differ between 2D/3D version as some pieces are not suitable for 2D (see Sec. 3.3)

*"reference image"* [17], thus limiting their potential applicability. In particular, since the early 2000's much of the literature focused almost exclusively on square jigsaw puzzles [82, 24, 97, 55, 4, 16, 64, 92, 72, 3, 5, 78, 67], then gradually expanding towards more general puzzles, such as puzzles with unknown piece orientation [30, 54, 73, 75, 93, 77, 76, 66], cases in which pieces from multiple puzzles were mixed together [30, 54, 58, 77, 76], missing pieces [30, 54, 58, 77, 76] noisy pictorial content [54, 7, 66, 75, 93, 76], gaps between neighboring pieces [61, 20], restricted deformations to the pieces' shapes [32], and puzzles of polygonal shapes [33, 17, 34]. In contrast, archaeological puzzles are considered unrestricted puzzles. Computational tools have revolutionized [31] this field by utilizing both the geometrical properties and pictorial content of pieces to match neighboring fragments [53, 81, 12, 28, 8, 71, 59, 62, 51, 6, 7, 48]. Among these works, Derech *et al.* [20] extrapolated the puzzle pieces content and attempted to resolve their spatial configuration using registration. Zhang *et al.* [96] employed an Internal Similarity Network (ISN) to score the dissimilarity of potentially neighboring fragments, after matching them by the similarity of their contour shapes. Zhang *et al.* [95] incorporated both the laminar textual content of pieces, all sampled from the Dunhuang scrolls, and their geometric shapes to align and match potential neighbors, while employing a hierarchical loops approach for global reconstruction. Recently, Cao *et al.* [11] utilized a multi-scale MobileViT classification network to evaluate the neighboring compatibility based on contour differences.

**Puzzle solving in 3D.** Puzzle solving in 3D can also be considered a reassembly problem with approaches such as furniture assembly [45]. However, these rely on the semantic properties (*i.e.* label or class) of the parts. In contrast, within the archaeological domain, frescoes have been extensively studied. [14, 29]. However, such approaches are stifled by limited data, which was addressed with the introduction of the *Breaking Bad* dataset [70]. In the synthetic *Breaking Bad*, the challenge is reconstructing a broken object from multiple artificially fractured fragments. However, those fragments do not have any semantics as in many real-world applications [9]. NSM [14] tried to address the two-part mating problem by emphasizing shape geometries over semantic information. *SE(3)-Equiv* [89] tackles the problem with specific design choices that go beyond object reassembly, *e.g.*, adversarial and reconstruction losses. [88] and [80] examine methods to efficiently plan physically plausible assembly motion and sequences for CAD based real-world assemblies by using physics-based simulations to efficiently explore a reduced search space. On the other hand, Jigsaw [50] is the first approach that tries to combine an approach using global and local hierarchical geometry features in order to match and align the fracture surfaces. Finally, DiffAssemble [67] is a Graph Neural Network (GNN) based architecture which learns to solve reassembly tasks using a diffusion model formulation.

## 3 Building the RePAIR dataset

The RePAIR dataset encompasses 1070 reconstructed 3D fragments ($\mathcal{F}$) with high resolution shape and texture, where $\mathcal{F} = \{f_i\}_{i=1,...,n}$. The fragments are organized into 117 groups ($\mathcal{G}$), as reassembled by the archaeologists, containing between 2 and up to 44 connected fragments per group. There is also a set of 150 (over the total 1070) ungrouped fragments, denoted as *"Isolated"*, for which their solution is still to be discovered, and supporting this way *open discovery* within the dataset. Following the description of the dataset's archaeological context in Section 3.1, we detail the digitization process in Section 3.2 and the generation of ground truth data in Section 3.3.

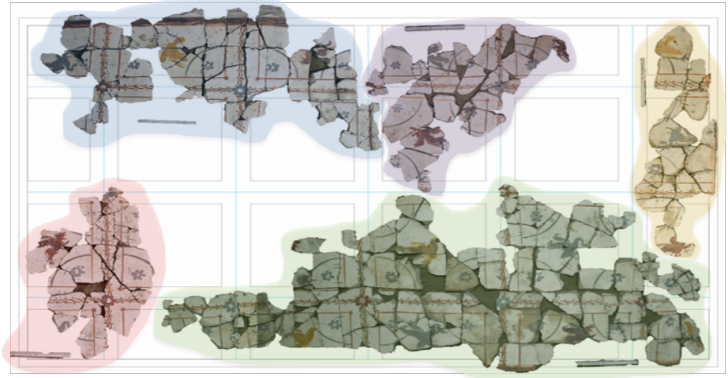

Figure 2: *Décor 1* is composed of five plaques, highlighted as islands of pieces in different colors. Each plaque is reconstructed by archaeologists through meticulous fragment connections and attribute analysis. (See Appendix E for more details).

## 3.1 Background archaeological activities at Pompeii park

The RePAIR dataset is a real-world example of years of archaeological fieldwork in the Pompeii Archaeological Park. It consists of fragments from several frescoes originally located in the *House of Painters at Work* which was destroyed both during the AD 79 eruption and by World War II bombings. The process of reassembling incomplete frescoes for archaeologists starts with painstaking cleaning and preparation of fragments. The assembly is then based on the analysis of several characteristics of the pieces, such as mortar layers and composition, traces of adhesion on the reverse side, smoothing stroke patterns of the plaster, and preparatory lines, alongside painted elements like background color and motif descriptions. Frescoes are assembled in a hierarchical manner starting from small clusters thanks to matching cues or comparison to similar patterns in related artworks and extending to larger clusters called "Plaques". Plaques are groups of attached fragments where no additional fragments can be added at that time, see Fig. 2.

The dataset contains more than a hundred groups solved by archaeological experts. Some of these groups can be connected to form an even larger part of a fresco. In fact, the fresco denoted as *Décor 1* has been studied and its schema identified, leading to a reassembly of large parts of it, see Fig. 2. Notably, not all fragments within *Décor 1* have been recovered and the final solution is still unknown, as the schematic solutions allow for more than one solution (see Appendix E).

The data acquisition process was conducted under the strict supervision of professional archaeologists, following best practices to ensure the safety of the fresco fragments. Our collaboration with the team onsite ensured a non-destructive process that staff was trained on the manipulation of ancient artifacts.

## 3.2 Digitizing Pompeii's fragments

Each fragment consists of $f = \{P, T\}$ with $P$ being a 3D point cloud $P \in \mathbb{R}^3$ and $T \in \mathbb{R}^2$ the corresponding texture coordinates. To acquire $P$, we used a Polyga H3 3D scanner [63], which captures geometric accurate models of the fragments, at $0.08$ mm accuracy, through a structured light sensor. Then, to acquire high-resolution textural content, we additionally used a Sony $\alpha$7c at 24.2-megapixel per image. To minimize environmental factors, including shadows and reflections, the scanner and camera were placed in a lighting box (as in Fig. 3) with the fragment placed on a turntable capturing $18$ viewpoints of the fragments. In addition, the top and bottom are captured and aligned. This dual configuration requires the registration of high-resolution images onto the geometric data by employing photogrammetric techniques to compute the texture $T$. We outline the key steps of the digitization pipeline in Figure 3 and metadata below, then refer the reader to the Appendix F for further details on the digitization process and metadata (Sec. F.1).

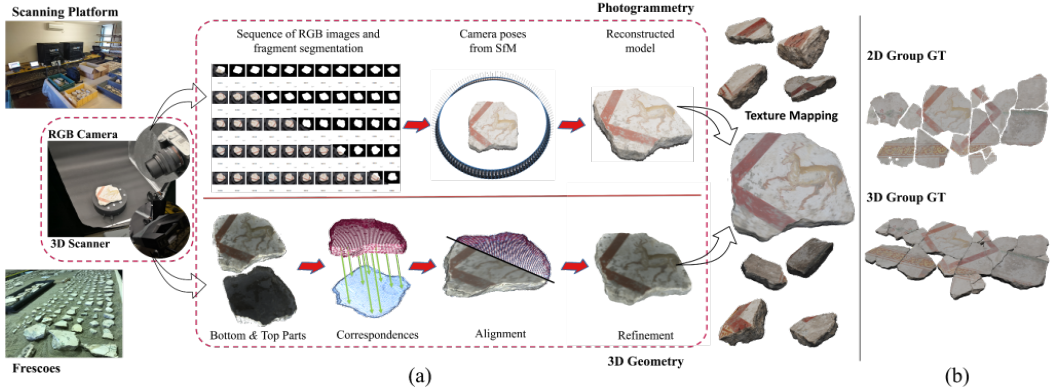

Figure 3: (a) The digitization process merges fragments 3D geometry from a structured light scanner and RGB images from a hi-res camera into high-detail textured 3D mesh. (b) The final groundtruth (GT) of fragments for 2D & 3D.

**Metadata.** For each fragment, the archaeologists provided artistic style and fresco family information. Moreover, geometric data (weight, fragment bounding box, *etc.*) and raw capture information and basic metadata (ID, version, *etc.*) have been included. The data is organized in *JSON* files for each fragment, adhering to the specifications that are commonly used in similar datasets ([70]).

### 3.3 Ground truth generation for 2D and 3D puzzle solving

Generating accurate ground truth is a challenge as the exact complete solution remains unknown even to experts in the archaeological field. Despite the unknown complete solution, archaeologists have physically assembled several subsets of pieces through years of fieldwork (See Fig. 2). These visual solutions serve as a reference for converting the archaeologists' findings into machine-readable 3D transformations. Following the group divisions outlined in Section 3, we create a Ground Truth (GT) solution in 2D and 3D for reassembling each group in our dataset (see Fig. 3).

**3D Ground Truth.** Generating 3D ground truth involves translating the physical reassembly by archaeological experts into 3D transformation matrices. We utilized Blender, to manipulate and align the 3D pieces. For each of the 117 groups in our dataset, we prepared a virtual 3D scene and manually reassembled them using reference images provided by the archaeologists. The 6DoF transformation matrices extracted from this process serve as the ground truth for the 3D reassembly task.

**2D Ground Truth.** Given the 3D puzzles ground truth, the 3D pieces were aligned with the textured surface's normal along the $Z$-axis, making the $XY$ plane a suitable approximation of the 2D image plane. Consequently, the 2D ground truth $(x, y, \theta)$ directly corresponds to scaled translations on the $X$ and $Y$ axes and the rotation angle (Euler) on the $Z$-axis. A scaling factor is applied to adjust the 3D millimeter coordinates to 2D pixel coordinates (see Appendix H).

***Note on 2D GT:*** *A small subset of the RePAIR fragments are located at the intersections of walls and ceilings, lacking a planar surface suitable for transformation into a 2D ground truth. Consequently, the 2D RePAIR dataset contains fewer fragments and puzzles compared to its 3D counterpart.*

## 4 Benchmarking RePAIR Dataset for Puzzle Solving

For both the 2D and 3D settings for puzzle solving, we constructed a set of benchmark experiments and common evaluation metrics in Section 4.1. We conducted a thorough quantitative assessment for 2D and 3D puzzle-solving in Section 4.2. For both scenarios, we define the set of baselines, their performance on RePAIR, and discuss the results. The RePAIR dataset is split into a provided training and testing with an $80/20$ percentage split, allowing the option of a validation set. More details on the evaluation metrics and the 2D and 3D methods are available in Appendix I.

### 4.1 Evaluation Metrics

Prior studies on 2D and 3D puzzle solving have introduced a range of evaluation metrics to assess errors related to geometry, such as fragment positioning, and neighbor consistency, which determines whether pieces are correctly aligned with their adjacent fragments.

**Evaluation metrics based on geometry.** Expanding Harel *et al.* [34] 2D metric to 3D, we define a $Q_{pos}$ metric, which scores the shared areas/volume between ground truth fragments pose (translation and rotation) and the solution given by the evaluated methods. It is important to make the metric invariant to rigid motion, avoiding scoring good solutions low only because they obtained the 'wrong' global rotations. Hence, we first apply a rigid transformation to the reconstruction on the largest fragment (named *anchor*) aligned in translation and rotation with the corresponding ground truth fragment. To calculate $Q_{pos}$, we first define a fragment's area or volume ($A(f_i)$). The shared area in 2D can be thought of as comparing the non-transparent pixels of two large canvases that contain all fragments in 2D. Similarly, in 3D, we compare the volume intersection of the registered point clouds. In addition, fragments are weighted based on their area/volume to reflect the importance of having an error on bigger fragments. The evaluation metric is formalized in the following measure:

$$Q_{pos} = \sum_{i=1}^{n} w_i \cdot \frac{\left| A(f_i \cap \tilde{f}'_i) \right|}{\left| A(\tilde{f}_i) \right|}, \text{ where } w_i = \frac{|A(f_i)|}{\sum_{k=1}^{n} |A(f_k)|} . \tag{1}$$

We also evaluate the $Root Mean Square Error (RMSE)$ for both translation in millimeters ($_{mm}$) and rotation in degrees($°$) computed relatively with respect to the ground truth, as:

$$\text{RMSE}(\mathcal{T}_{mm}) = \frac{1}{\sqrt{n}} \sum_{i=1}^{n} \|\hat{t}_i - t_i^{gt}\|_2, \quad \text{RMSE}(\mathcal{R}°) = \frac{1}{\sqrt{n}} \sum_{i=1}^{n} \|\hat{R}_i - R_i^{gt}\|_2. \tag{2}$$

where $\hat{t}_i$ is the predicted translation, $t_i^{gt}$ is the ground truth translation, $\hat{R}_i$ is the predicted rotation, and $R_i^{gt}$ is the ground truth rotation of the $i$-th fracture. Notice that the $Q_{pos}$ score metric is indirectly influenced by the RMSE scores. Even slight rotations, which may alter a fragment's overall position in the workspace, or slight translations in the opposite direction of its ground truth position, can significantly impact its $Q_{pos}$ score.

**Evaluation metrics based on neighbors consistency.** These metrics evaluate the consistency of neighbors given a ground truth mating graph, *i.e.* a graph that indicates which are the neighbors of a specific fragment. Intuitively, we define neighboring fragments in 2D or 3D space as two pieces that are close to one another in the assembly (either the ground truth or the solver's one) and not separated by a third fragment. To determine the neighbors of each fragment, we established a threshold representing the maximum distance observable in RePAIR between adjacent pieces. Therefore, fractures within this threshold are deemed neighbors, while those beyond it are not considered as such. By comparing edges of a solution's mating graph $M_{sol}$ to the ground-truth ones $M_{GT}$, we define several evaluation metrics, as proposed in [33, 34] for their crossing cuts polygonal model:

$$Q_{\text{precision}} = \frac{\sum_{\{p_i,p_j\} \in M_{sol} \cap M_{GT}} (|A(p_i)| + |A(p_j)|)}{\sum_{\{p_i,p_j\} \in M_{GT}} (|A(p_i)| + |A(p_j)|)}, \quad Q_{\text{recall}} = \frac{\sum_{\{p_i,p_j\} \in M_{sol} \cap M_{GT}} (|A(p_i)| + |A(p_j)|)}{\sum_{\{p_i,p_j\} \in M_{sol}} (|A(p_i)| + |A(p_j)|)} \tag{3}$$

$Q_{\text{precision}}$ counts true mattings by the solver, while $Q_{\text{recall}}$ represents the fraction of predicted neighbors that are true neighbors. Both metrics are weighted by fragments' area/volume, giving greater importance to larger pieces of the puzzle. They can be consolidated with their harmonic mean into the $F1$ score, to quantify performance: $F1 = 2 \frac{Q_{\text{precision}} \cdot Q_{\text{recall}}}{Q_{\text{precision}} + Q_{\text{recall}}}$

### 4.2 2D puzzle solving evaluation

We consider only SoTA approaches able to reconstruct archaeological puzzles, where fragments are of unrestricted shapes, often eroded (or more generally, deformed geometrically), and possibly pictorial to some degree. Very few examples are not bespoke to data or task and very few have code openly available. We evaluate three different baseline approaches, one from literature and two naive baselines, examining both their ability to recapture the fragments neighborhood graph, (*a.k.a.* the *mating* graph) and the geometric quality of reconstruction.

Table 3: Results on RePAIR 2D Puzzle solving.

| Method | $Q_{pos}$ ↑ | RMSE ($\mathcal{R}°$) ↓ | RMSE ($\mathcal{T}_{mm}$) ↓ | Precision ↑ | Recall ↑ | F1 ↑ |
|---|---|---|---|---|---|---|
| Derech *et al*. [20] | 0.037 | 80.964 | 139.495 | 0.454 | 0.527 | 0.471 |
| Genetic Optimization | 0.047 | 85.625 | 151.714 | 0.313 | 0.662 | 0.394 |
| Greedy Geom Match | 0.023 | 76.987 | 135.946 | 0.297 | 0.470 | 0.351 |

**Baseline Methods.** We benchmark the performance of three SoTA solvers on our dataset: *i)* Derech *et al*. [20] Archaeological Puzzle Solver applies a greedy next best piece algorithm based on the texture. However, they use an outdated extrapolation process, which we replaced with the stable-diffusion extrapolation process [34]. *ii)* Genetic Algorithm reconstruction uses a fitness function of the geometry, using the area of the puzzle's bounding rectangle and the intersection area of overlapping pieces. Since perfect solutions tend to minimize both of these values. Naturally, this does not guarantee a good solution. *iii)* Greedy geometric matching, from a random seed fragment, iteratively extends the fragment pose in a greedy fashion based on geometry in contrast to [20], which is based on texture compatibility, see Apendix I.2 for more details.

**Results discussion.** We show the results of the 2D baseline methods in Table. 3, experimenting on the test set (excluding one group, Sec. 3.3). Although all of these methods achieve relatively poor results, we note that the Genetic Optimization and the Greedy Geometric Matching approaches, both proposed in this paper, achieve competitive results to those achieved by Derech *et al*. [20], even surpassing them in some metrics. Overall, however, existing methods do not seem to cope well with realistic (archeological) puzzle solving, and leave much room for original research on solving challenges like the dataset presented in this paper. For qualitative results, see Fig. 4.

### 4.3 3D Puzzle Solving Evaluation

For 3D puzzle solving, we evaluate six methods using the same metrics as in the previous section.

**Baseline Methods.** We benchmark the performance of five SoTA learning-based shape assembly methods on our dataset: Global [49], LSTM [90], DGL [94], SE(3)-Equiv. [89], and DiffAssem-ble [67] (see Appendix I.3 for more details). While NSM [14] is also a learning-based method designed for geometric shape assembly, it focuses solely on assembling two parts, making it unsuitable for the multi-part assembly task in this paper. Furthermore, we did not benchmark RGL-NET [35] because it relies on predefined part ordering for shape assembly. Although part ordering is logical in semantic assembly, it's not as clear in geometric assembly. Without part ordering information, RGL-NET effectively behaves like DGL. As a result, NSM and RGL-NET were excluded from our performance benchmarking.

Table 4: Results on RePAIR 3D Puzzle solving.

| Method | $Q_{pos}$ ↑ | RMSE ($\mathcal{R}°$) ↓ | RMSE ($\mathcal{T}_{mm}$) ↓ | Precision ↑ | Recall ↑ | F1 ↑ |
|---|---|---|---|---|---|---|
| Global [49] | 0.127 | 88.606 | 58.020 | 0.707 | 0.712 | 0.683 |
| LSTM [90] | 0.071 | 89.071 | 61.732 | 0.492 | 0.880 | 0.585 |
| DGL [94] | 0.119 | 87.611 | 58.387 | 0.669 | 0.751 | 0.659 |
| SE(3)-Equiv. [89] | 0.093 | 70.816 | 56.046 | 0.724 | 0.670 | 0.681 |
| DiffAssemble [67] | 0.099 | 69.536 | 67.960 | 0.506 | 0.905 | 0.599 |

Table 5: Generalization Results on 3D puzzle solving (trained on RePAIR and tested on Breaking Bad [70] dataset).

| Method | $Q_{pos}$ ↑ | RMSE ($\mathcal{R}°$) ↓ | RMSE ($\mathcal{T}_{mm}$) ↓ | Precision ↑ | Recall ↑ | F1 ↑ |
|---|---|---|---|---|---|---|
| Global | 0.291 | 85.506 | 74.816 | 0.923 | 0.894 | 0.890 |
| LSTM | 0.245 | 86.733 | 73.295 | 0.774 | 0.871 | 0.777 |
| DGL | 0.296 | 84.634 | 63.378 | 0.949 | 0.917 | 0.921 |
| SE(3)-Equiv. | 0.220 | 83.824 | 72.358 | 0.893 | 0.886 | 0.859 |
| DiffAssemble | 0.174 | 85.913 | 85.929 | 0.654 | 0.879 | 0.692 |

Table 6: Generalization Results on 3D puzzle solving (trained on Breaking Bad [70] dataset and tested on RePAIR).

| Method | $Q_{pos}$ ↑ | RMSE ($\mathcal{R}°$) ↓ | RMSE ($\mathcal{T}_{mm}$) ↓ | Precision ↑ | Recall ↑ | F1 ↑ |
|---|---|---|---|---|---|---|
| Global | 0.047 | 84.560 | 64.509 | 0.569 | 0.725 | 0.606 |
| LSTM | 0.042 | 91.187 | 67.103 | 0.541 | 0.838 | 0.629 |
| DGL | 0.063 | 84.532 | 67.723 | 0.543 | 0.829 | 0.599 |
| SE(3)-Equiv. | 0.063 | 86.990 | 59.247 | 0.668 | 0.770 | 0.688 |
| DiffAssemble | 0.066 | 85.055 | 69.233 | 0.503 | 0.842 | 0.587 |

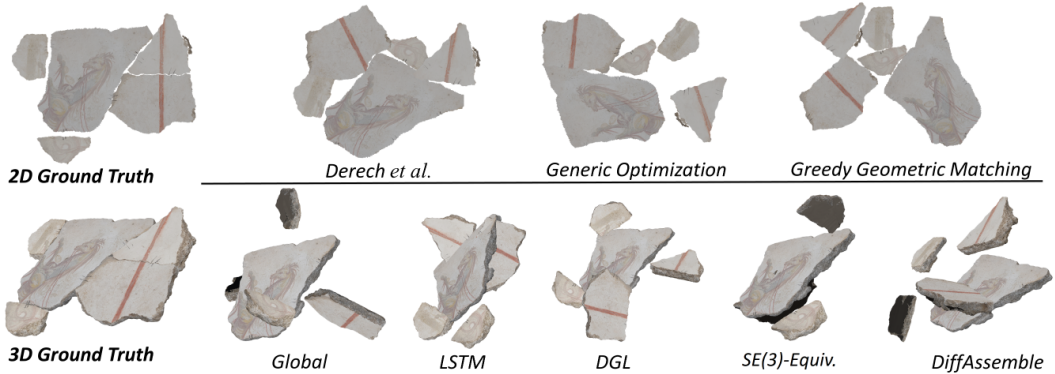

Figure 4: 2D and 3D qualitative results in one of the RePAIR dataset groups (for additional qualitative results please refer to Appendix I).

**Results discussion.** We report the results of testing the baseline models for 3D assembly in Table 4 where it is evident that no single method outperforms all others across the different metrics. It is noteworthy that, akin to 2D methods, 3D approaches also struggle to accurately identify neighboring fragments, as evidenced by the F1 scores. Specifically, Diffassemble and LSTM achieve impressive recall values of 0.905 and 0.880, respectively, but these are offset by low precision values of 0.506 and 0.492. In contrast, the other three methods demonstrate more consistent precision and recall results. Our findings highlight a considerable gap in solving this problem. While fractures in the Breaking Bad dataset are easily identifiable and attributable to object components, our dataset presents substantial challenges in interpreting fracture semantics. This issue is further compounded by the baseline models' lack of color consideration for each fracture. Qualitative results are shown in Fig. 4.

While our dataset being smaller compared to "Breaking Bad," we believe that methods trained on this dataset will generalize reasonably well within the same domain of objects, such as other frescoes in Pompeii or artifacts from similar heritage sites. However, generalization to datasets outside of this domain, particularly those with artificial or dissimilar characteristics, may be more challenging. To address this concern, we have conducted an experiment to evaluate the generalization capabilities of baseline approaches. Tables 5 and 6 show how the benchmark baselines perform when trained in our RePAIR dataset and tested on a Breaking Bad dataset subset (2000 objects) and when trained on the Breaking Bad dataset and tested on our RePAIR dataset respectively. The tables show that the methods generally perform better when trained on RePAIR and tested on baselines (Tab. 5) than the opposite (Tab. 6). This is evident from higher scores across most metrics in Table 5. The lower scores in Table 6 suggest that generalization from other datasets to RePAIR is more challenging than generalizing from RePAIR to other datasets. For example, for the $Q_{pos}$ metric the higher value of 0.066 in Table 6, is significantly lower compared to the corresponding higher value for the same metric, *i.e.* 0.296, in Table 5. The same behavior is noticed in the metrics of $Q_{precision}$, $Q_{recall}$ and $F1$. On the other hand, the $RMSE$ reported errors both for the rotation and translation seem to follow a more balanced behavior with the errors to be closer to each other for each case with only $\approx$1 degree and $\approx$4mm error respectively. These findings show that the RePAIR dataset's unique complexities are not easily captured by standard artificial datasets, making it valuable for developing more robust models for the assembly task.

## 5 Conclusion and Limitations: The way ahead

This paper proposes the RePAIR dataset, a complex real-world puzzle-solving dataset, where we demonstrated the challenge in both 2D and 3D for the current SoTA and naive baseline methods. Real world datasets are inherently of higher complexity than synthetic datasets, as breaking patterns, erosion and deterioration are realistic. In 2D, the SoTA lacks methods that are able to handle such irregular and eroded pieces, as most of the literature addresses synthetic setups (*e.g.*, square puzzles) that limit their real-world applicability. 3D puzzle-solving methods for irregular and eroded do exist, but their performances are far from solving the problem even on relatively small challenges. Both 2D and 3D puzzle solving methods exhibit significant sensitivity and large discrepancies in their evaluation metrics such as absolute position ($RMSE$) or neighbor metrics ($Q_{\text{precision}}$, $Q_{\text{recall}}$, $F1$). Furthermore, there is surprisingly little difference in performance between 2D and 3D results, indicating that the problem is similarly difficult for both sets of methods. This is especially surprising for Derech *et al*. [19], the only method that handles texture information, which is crucial during human groundtruth generation for both archaeologists and the translation to 6DoF pose.

The current performance of SoTA methods clearly indicates that many challenges in realistic puzzle solving still lie ahead, regardless of the dimension (2D or 3D) where it is considered. Most prominent are the ability to exploit degraded geometry (as is the case in eroded fragments), coping with poor or scarce pictorial information (as in the RePAIR fresco), and the inherent combinatorial complexity. Despite many years of research on 2D pictorial puzzle solving, few methods are able to handle irregular shape pieces, and even fewer when such fragments are not rich pictorially. 3D methods lack focus on geometrically degraded fragments, and most often ignore pictorial information altogether. We recognize that the current baseline approaches are not specifically designed to address these more complex, real-world breaking patterns. We believe that this complexity is a key strength of our dataset, as it provides a more accurate representation of real-world scenarios. Thus, we hope by introducing the RePAIR dataset will help and motivate the community to have a standard baseline to develop methods that can be applied to real-world puzzle solving methods, either for data-driven approaches (via training) and/or for evaluation. Moreover, as a real-life dataset, emerging from a real archaeological origin, the RePAIR dataset is designed to foster the development and present an opportunity to incorporate and exploit additional cues and features emerging from art history and archaeology knowledge. In practice, this suggests the need for more advanced methods that can better handle these challenges and possibility to endow the current frontiers of geometrical and pictorial tools in puzzle solving with novel methods from language and knowledge inference (*e.g.*, via Large Language Models, LLMs, or Multimodal Large Language Models, MLLMs), all may prove vital to handle the complexities of real-life puzzles like the RePAIR dataset.

Another interesting aspect would be to explore whether 3D puzzle-solving can be considered and solved as a shape-repair (or completion) problem. Relevant works in the literature involve filling in missing parts of a single, incomplete object. These methods are designed to infer and generate the missing geometry of a known object based on existing data. They often use occupancy functions [41, 42] or generative adversarial networks [37] to predict the missing portion of the fractured piece. This problem is different from reassembly, as in the proposed problem, we have multiple (known) fragments that need to be matched and registered to obtain a consistent 3D shape, as provided by the ground truth. However, we believe shape repair could be a complementary task to the reassembly task and also be very useful for archaeologists as: a) It can repair eroded surfaces for each fragment, so to provide a better alignment and a more consistent overall surface; b) It can complete the puzzle with missing pieces that have been lost or irremediably destroyed during the collapse.

Finally, we would like to conclude with some societal concerns that might be raised, especially with the blooming of AI. This could be the potential unethical uses of this dataset for artifact forgery. The approach certainly does not preclude such a case from happening, however the consensus on Open Access Heritage acknowledges that the benefits outweigh the risks, particularly in light of recent instances of destruction and theft. While AI tools for reassembly could theoretically be misused for criminal purposes, their potential to reconstruct thousands of frescoes or other objects from heritage sites across the globe offers significant insights into historical societies and cultures. It is recognized that access to digital twins of historical fragments may raise ethical concerns, especially with the potential for future 3D printing in comparable materials or the distribution of digital versions. However, to date, there is no agreement upon 3D watermarking approach [56].

## 6 Acknowledgements

This work is part of the RePAIR project that has received funding from the European Union's Horizon 2020 research and innovation programme under the grant agreement No. 964854.

## Overview

We provide additional details, results, and material to complement the main paper. Specifically, we include the following additional information:

- Broader impact (Appendix A).
- Dataset access and storage (Appendix B).
- Source code (Appendix C
- Licensing (Appendix D).
- Archaeological background for the RePAIR Dataset (Appendix E).
- Additional details on the digitization procedure (Appendix F).
- Rendering the associated 2D image of each 3D fragment (Appendix G).
- Ground truth generation (Appendix H).
- Additional experimental details (Appendix I).

## A Broader impact

Fracture reassembly is crucial in real-world scenarios, such as recovering fragmented artifacts or broken objects. Recently, machine learning algorithms trained on large-scale datasets have significantly advanced this task. However, extensive experiments and results reveal that fractured shape reassembly is still an open problem. We believe our RePAIR dataset and benchmark represent a key step towards teaching machines to reassemble real-world fractured objects. Our dataset will support future research in this field as well as in fragment registration and reconstruction methods and ultimately enable robots to relieve humans from the task of assembly.

**Potential negative societal impacts.** Apart from the discussion in Section 5 regarding unethical uses of our dataset, we do not anticipate any other significant risks of human rights violations or security threats arising from our dataset and its potential applications. However, as our dataset contributes to the field of fracture reassembly, it could raise concerns about assembly algorithms. Despite advances in learning methods, human trust in AI remains an issue. For instance, professionals in archaeology or cultural heritage might be skeptical of AI assistance with delicate objects. Therefore, trained algorithms should be used under supervision and cannot fully replace human expertise. Overall, we view the technical outcomes of this paper as requiring human cooperation to avoid negative societal impacts.

## B Dataset access and storage

We provide a compressed version of our RePAIR dataset in the Zenodo open repository, in the following link: `https://zenodo.org/records/13993089`.

The project page can be found at `https://repairproject.github.io/RePAIR_dataset/` where more information, instructions and details together with user-friendly gallery views of parts of the dataset are available.

**Storage and Formats** Our full dataset contains over one thousand individual fractured fragments divided into groups with its corresponding folder. Considering the 3D dataset, each fragment is saved as a mesh using the widely `.OBJ` format with the corresponding material (`.MTL`) and texture (`.PNG`) file. The meshes are already in the *assembled* position and orientation, so that no additional information is needed. Regarding the 2D dataset, each fragment is saved as a `.PNG` image and each group has the corresponding ground truth transformation to solve the puzzle as a `.TXT` file. All additional metadata information are given as `.JSON` files.

## C  Source code

The code is organized under the RePAIR organization in Github `https://github.com/RePAIRProject`. Since the creation of the dataset involved several steps, we have different sub-repositories covering each part of the work.

- **Fragment Digitization**
  (`https://github.com/RePAIRProject/fragment_digitization` repository)
  It contains the scripts related to the pipeline for the fragments digitization (3D reconstruction, texture mapping, partial meshes alignment).
- **3D Segmentation**
  (`https://github.com/RePAIRProject/segment_3Dfragment` repository)
  It contains the code for segmenting the 3D fragments into the painted surface, the opposite one and the sidewalls.
- **Rendering (from 3D to 2D)**
  (`https://github.com/RePAIRProject/pcd2images` repository)
  It contains the code and the workflow to render the 2D images from the 3D fragments (best results when used after 3D segmentation).
- **Ground Truth preparation**
  (`https://github.com/RePAIRProject/repair_ground_truth` repository)
  It contains the guide for the preparation of the virtual scenes and the steps to follow to do the manual reassembly (using Blender) for the ground truth generation.
- **2D Puzzle Solving**
  (`https://github.com/RePAIRProject/2D-baselines` repository)
  It contains the code to evaluate the baseline approaches for 2D puzzle solving.
- **3D Puzzle Solving**
  (`https://github.com/RePAIRProject/3D-baselines` repository)
  It contains the code to evaluate the baseline approaches for 3D puzzle solving.

## D  Licensing

We release each model in our dataset under a custom license which means that the data can be downloaded, reviewed and reused under two conditions: attribution and non-commercial purposes. Moreover, the authors accept full responsibility for any rights violations and confirm the licenses associated with the dataset and code.

## E  Archaeological background for the RePAIR dataset

As a prominent example, the RePAIR addresses the reassembly process of broken ceiling frescoes and fragments from the *House of Painters at Work* located at the Archaeological Site of Pompeii. This house was initially discovered in 1912 by Vittorio Spinazzola [84] and subsequently excavated by Antonio Varone [85] in 1987. It features a large garden encircled by painted rooms, with evidence suggesting that construction work was in progress at the time of the AD 79 eruption, likely to repair damage caused by the AD 62 earthquake. One substantial room displays signs of ongoing fresco redecoration, which has led to the house's modern name. The building sustained additional damage during World War II bombings, which heavily affected many parts of the ancient city. All these fragments are currently stored in the storerooms of the *Casina Rustica* within the Archaeological Site of Pompeii.

The classical archaeological method for reconstructing fragmented and incomplete frescoes involves multiple stages, different from those used for in-situ decorations [86]. Initially, after excavation, the fragments are cleaned and inventoried according to excavation data (Stages 1 and 2). Fragments must be kept as found, without sorting by motif or color, to preserve vital connections. Roman wall painting specialists analyze the fragments by cataloging features such as mortar layers and composition, traces of adhesion on the reverse side, smoothing stroke patterns of the plaster, and preparatory lines, alongside painted elements like background color and motif descriptions. This examination of all sides (obverse, reverse, and edges) of the fragmented plaster guides the search for connections. This process begins with individual fragment groups and advances through the excavation's stratigraphic unit (Stage 3), gradually reassembling the elements in a sandbox (Stage 4, Fig. 5). Throughout the process, comparisons with other paintings are made to help understand incomplete patterns. These comparisons serve as a basis for justifying the proposed restoration, which is documented through surveys and photographs of the reconstituted plates in the sandbox (Stage 5, Fig. 6), followed by restitution and the preparation of a final report (Stages 6 and 7).

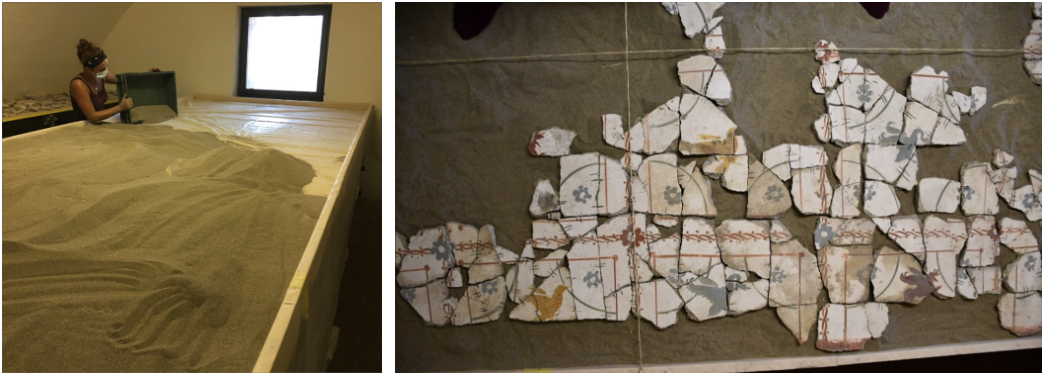

Figure 5: Reassembly of the fragments in a sandbox by the archaeologists.

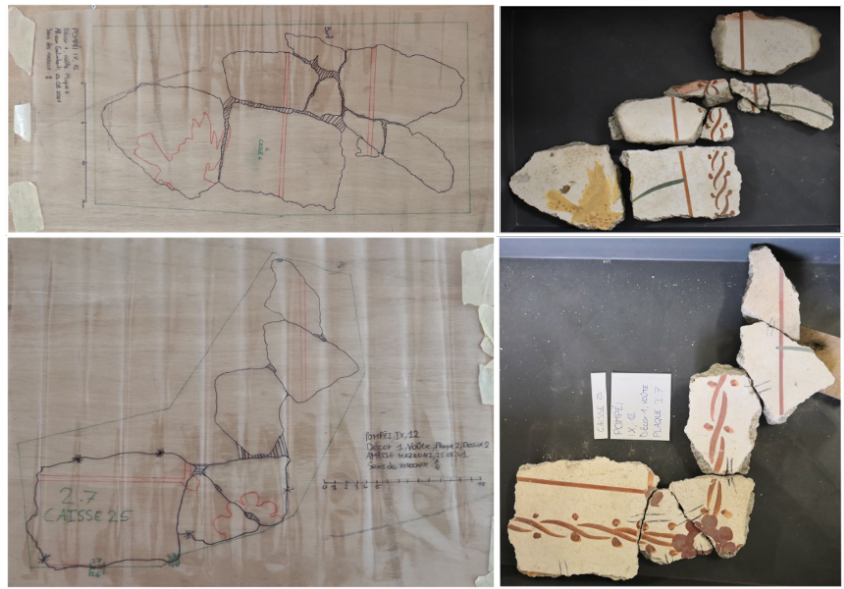

Figure 6: Graphic documentation examples of reassembled fragments by the archaeologists.

A substantial subset of the dataset pertains to what archaeologists have termed *Décor 1*, representing a significant portion of a fresco whose solution is partially known through manual alignment efforts. This subset serves as a valuable reference for training and evaluating algorithms pertaining to puzzle-solving, visual enhancement, fragment restoration, and handling of missing fragments and erosion. Not all fragments within *Décor 1* have been recovered. Archaeologists proposed two solutions for *Décor 1*'s reconstruction based on what they call "Plaques", shown in Figure 2. The arrangement of the figured motifs (C, O, G) is attested on these five reconstructed plaques. The cohesive arrangement of figurative motifs is repeated diagonally and alternate on each line. The exact position of the two plaques was uncertain as they could be interchanged, affecting the alternation of the patterns. The drawing enabled schematization of the two potential options Version 1 and Version 2 as these depicted in Figure 7. Each places two unconnected plaques in different spots, resulting in these two most probable distinct configurations. In the first version, the oblique alternation of the animals is preserved along one long side of the vault, with the pattern reversed on the opposite side. This creates a sequence of griffins followed by birds, with a swan in the corner. In the second version, the

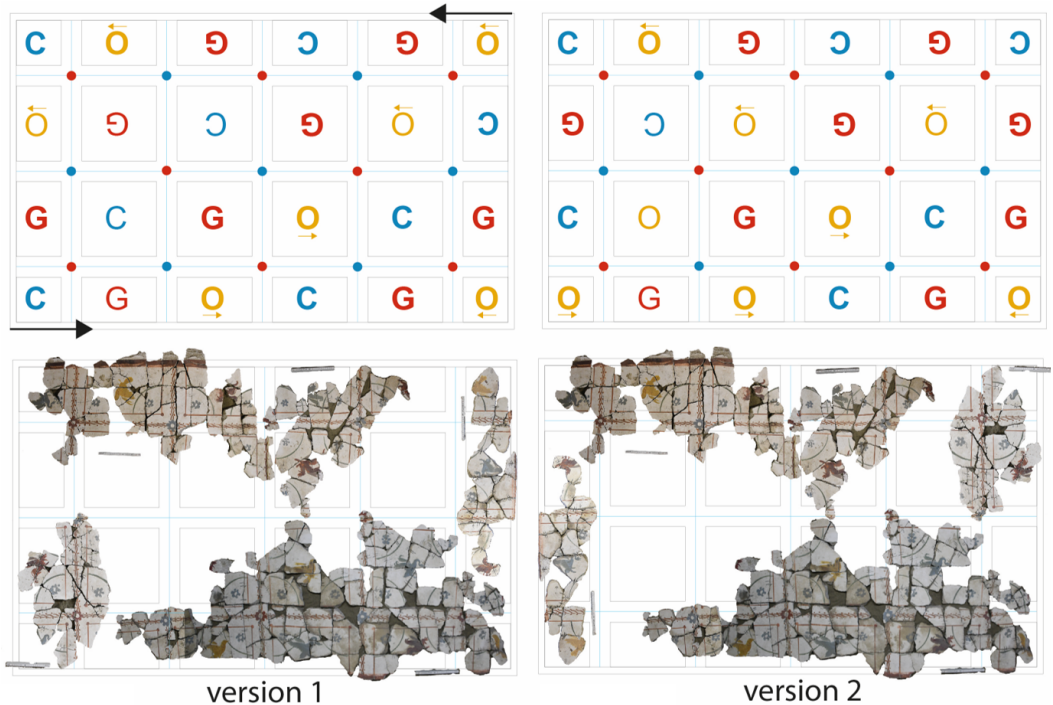

version 1                                          version 2

Figure 7: *Décor 1*, Version 1 and 2 as reconstructed by archaeologists. The figurative motifs and patterns used to understand the system of the repeat-pattern decoration are illustrated on the top row. Bottom row illustrates the fragments as these are separated in 5 different plaques and the guessing order of each.

organization becomes more complex and loses the logical pattern of oblique alternation, resulting in a swirling effect that doesn't match the typical regularity of Roman repeat patterns. Therefore, the archaeologists opted for the first version for the final rendering. The colored letters (C, O, G) in the motifs represent animals: C (blue) for "Cygne bleu" (blue Swan), O (yellow) for "Oiseau jaune" (yellow Bird), and G (red) for "Griffon rouge" (red Griffin). Their placement and arrows indicate the animals' facing direction.

The RePAIR dataset contains also a subset of 3 test frescoes artificially made, from mortar, and broken in the style of the ancient frescoes. Although these fragments are not as old as the real ones, the large number of matches helps us test whether we can effectively reconstruct an entire fresco since it features similar characteristics. Another portion of the dataset comprises fragments for which a reference solution has yet to be identified, we refer to them as *"Isolated"* (Fig. 8a). While archaeologists anticipate these fragments should fit together, but the precise solution is not guaranteed and therefore still remains elusive and supporting an *open discovery* component.

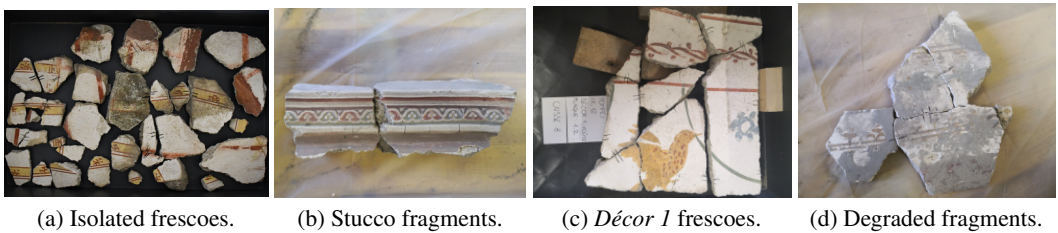

(a) Isolated frescoes.    (b) Stucco fragments.    (c) *Décor 1* frescoes.    (d) Degraded fragments.

Figure 8: Illustration of the RePAIR dataset fragments variety.

Overall, the RePAIR dataset is encompassing fragments of varying size, shape, color, and characteristics. These include *Stucco* fragments (angular pieces lacking a textured flat surface), fragments with distinct patterns compared to *Décor 1*, and fragments exhibiting varying degrees of degradation, see Fig. 8.

Ensuring comprehensive surface coverage while maintaining the artifacts' archaeological characteristics is of paramount importance. Furthermore, to ensure the efficacy of registration algorithms, the input data must adhere to stringent accuracy criteria encompassing uniform brightness, scale consistency, and geometric precision. As mentioned, the Polyga H3 3D scanner [63] is utilized for acquisition, leveraging its two integrated external cameras and the projector to capture geometric fidelity under controlled lighting conditions. Its capability to record point clouds at a resolution up to 80 microns (0.08 mm) facilitates high-geometric data accuracy capture. Subsequently, the raw 3D data undergoes processing to generate up to 1.5 million points per second.

# F   Additional details on the digitization procedure

Conversely, the cameras of the device do not ensure high-resolution imagery, exhibiting substandard quality in terms of both color accuracy and pictorial details. To address this limitation, the reconstruction process is augmented by integrating the Sony $\alpha$7c high-resolution mirrorless digital camera. This camera boasts a 24.2-megapixel optical sensor and is capable of capturing images with resolutions of up to 6000x3376 pixels. This represents a significant enhancement compared to the 3-megapixel sensor and 720x480 pixels resolution of the images that can be obtained with the Polyga scanner.

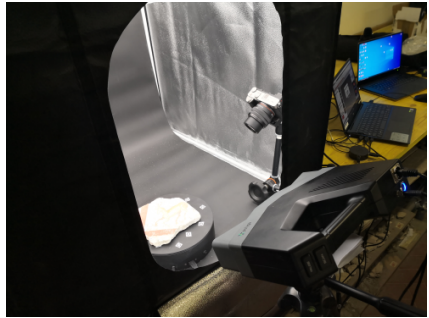

Figure 9: The scanning setup.

The usage of a lighting box ensures consistent isometric ambient lighting conditions, mitigating the impact of any potentially disruptive external light sources. By eliminating problematic factors such as shadows or reflections, the final quality of the captured images is preserved and upheld. The complete setup was consisting of a lighting box, one turntable, a high resolution camera and a 3D scanner, see Figure 9.

The scanned fragments exhibit diverse characteristics, enabling a multitude of analyses. Significant attention was devoted to selecting the subset of fragments for scanning, allowing concurrent tasks to commence while scanning operations persist.

## F.1   Additional technical details of the frescoes digitization pipeline

To acquire the ground truth digitized models, each scanned piece has undergone a series of post-processing procedures, which are detailed below:

**Align and merge top and bottom parts into one piece.** For each fragment $f_i$, we capture top and bottom independently, which was performed automatically employing a Truncated Least Squares (TLS) registration [91]. The alignment process involves extracting robust putative correspondences $(\mathbf{a}_i, \mathbf{b}_i), i = 1, ..., N$ and determining the optimal transformation to align the top, $\mathbf{a}$, and bottom, $\mathbf{b}$ point clouds.

$$\min_{s>0, R \in SO(3), \mathbf{t} \in \mathbb{R}^3} \sum_{i=1}^{N} min \left( \frac{1}{\beta_i^2} \|\mathbf{b}_i - sR\mathbf{a}_i - \mathbf{t}\|^2, \bar{c}^2 \right) \tag{4}$$

Where $s$, $R$, and $t$ are the unknown (to-be-computed) scale, rotation, and translation respectively and which are computed through a least-squares solution of measurements with small residuals $(\frac{1}{\beta_i^2} \|\mathbf{b}_i - sR\mathbf{a}_i - \mathbf{t}\|^2 \leq \bar{c}^2)$ and a given noise bound $\beta_i$. The (RANSAC) [25] algorithm is then leveraged to handle outlier correspondences, ensuring the determination of the best transformation parameters.

**Mesh refinement of the complete pieces.** The alignment process may not ensure a perfect merge between the two parts, especially for complex and irregular shapes like broken surfaces (see Fig. 10). Therefore, to address this issue as well as any nuisances on the mesh surface, *e.g.* holes, overlapping faces, *etc.* and achieve a seamless mesh, a mesh refinement step is integrated into the pipeline based on the Screened Poisson surface reconstruction algorithm [39] where the surface is computed based on an indicator scalar function $\chi$ whose gradient best approximates a vector field $\vec{V}$ defined by sample points, $min_\chi \left\| \nabla_\chi - \vec{V} \right\|$.

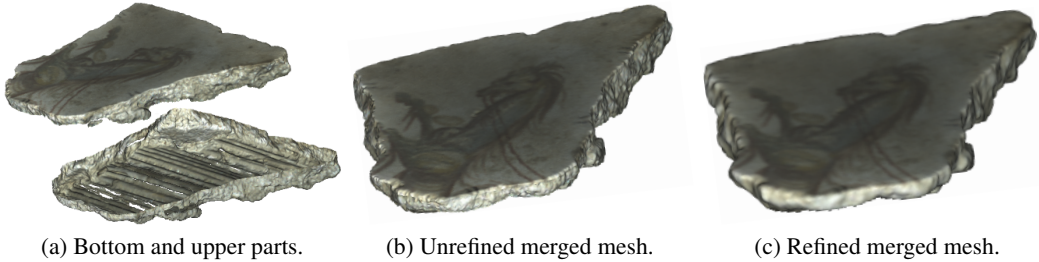

(a) Bottom and upper parts.  (b) Unrefined merged mesh.  (c) Refined merged mesh.

Figure 10: Illustration of the sub-parts merging and refining procedures. Notice the difference on the fractured area of the pieces between the (b) non-refined, and (c) refined versions.

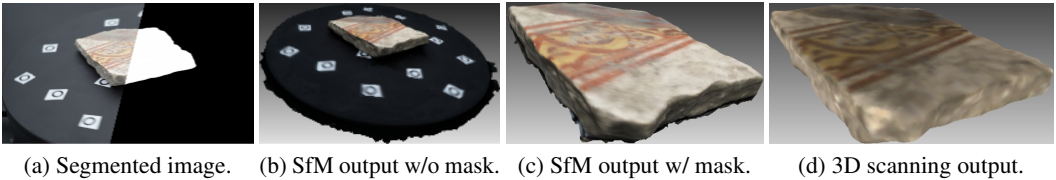

(a) Segmented image.  (b) SfM output w/o mask.  (c) SfM output w/ mask.  (d) 3D scanning output.

Figure 11: Illustration of mask segmentation sample.

**Create high-resolution texture maps using photogrammetry.** We utilize the high-resolution images from the digital camera to generate a secondary 3D model, facilitating the mapping of high-resolution texture information onto the 3D model reconstructed from the 3D scanner. For this purpose, we employ Structure from Motion (SfM) [83, 68].

$$\min_{\{\mathbf{P}_j\},\{C_i\}} \sum_{i \sim j} \left( x_{i,j} - \frac{C_{i1}^T \mathbf{P}_j}{C_{i3}^T \mathbf{P}_j} \right)^2 + \left( y_{i,j} - \frac{C_{i2}^T \mathbf{P}_j}{C_{i3}^T \mathbf{P}_j} \right)^2 \tag{5}$$

where, $(x_{ij}, y_{ij}) \in \mathbb{R}^2$ denote the computed projection of point $\mathbf{P}_j$ (in homogeneous coordinates $[\mathbf{P}_j, 1] \in \mathbb{R}^4$) onto the image plane of the camera $C_i$, with $C_{ik} \in \mathbb{R}^4$ denoting the $k$'th row of the $C_i (1 \leq k \leq 3)3 \times 4$ camera matrix, and $i \sim j$ indicating that the $j$'th scene point is visible by the $i$'th camera. To achieve our goal of reconstructing only the fragments and not the entire scanning scene, using segmentation masks allows us to avoid reconstructing the background environment surrounding each fragment during the scanning process. We adopted a semi-supervised learning approach, comprising three steps: *i) Initial user annotation*, where a user annotates the first image for each fragment using a guided interactive segmentation scheme [74], requiring minimal user input to generate a segmentation mask (supervised part) *ii) Temporal consistency extension*, where the initial mask is extended to subsequent images by leveraging temporal consistency between consecutive frames [15], exploiting the limited degrees of rotation between each frame and *iii) Network fine-tuning*, where annotated images are utilized to fine-tune a pre-trained network for segmentation, ensuring accurate segmentation, even along fragment borders. We chose to build upon Mask R-CNN [36], a robust state-of-the-art method capable of detecting and segmenting objects in the scene.

Through segmentation, we create a model with characteristics akin to those generated by the Polyga 3D scanner facilitating comparison between the two models and enabling texture mapping transfer, see Figure 11.

**High-resolution texture mapping to the refined pieces.** Upon reconstructing the second 3D model via the photogrammetry pipeline, we proceed to transfer its texture onto the model obtained from the 3D scanner. This is done by utilizing a rasterization pipeline [10]. This mapping involves associating the collected high resolution images and their extracted camera poses from SfM with the 3D scanned model. The rasterization mode operates by perspective-projecting the color information from all active color rasters onto the 3D mesh, creating texture $T$.

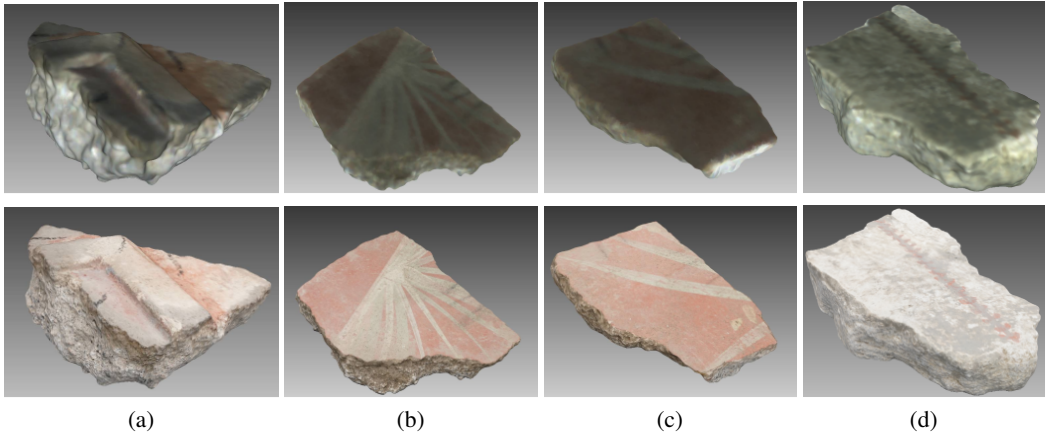

|   (a)   |   (b)   |   (c)   |   (d)   |

Figure 12: Illustration of the low (upper row) and high (bottom row) resolution outputs for samples in the RePAIR dataset.

Samples showcasing the final result of low- and high-resolution texture mapping over the 3D scanned models are depicted in Figure 12.

**Ground truth metrics and metadata.** As we have noted the metadata information is organized and stored in individual *JSON* files for each fragment, adhering to the specifications that are commonly used in similar datasets already existing in the literature. Each metadata JSON file shares the same name as the corresponding fragment it pertains to and comprises the following key attributes:

- **Acquisition Date:** Indicates the date when the corresponding fragments underwent scanning.
- **Artistic Style\*:** Specifies the specific artistic style to which the piece belongs to.
- **Filename(s):** Encompasses the filename(s) of the 3D mesh representing the fragment, typically stored as *.ply* and/or *.obj* files.
- **Fresco Family\*:** Identifies the fresco family to which the fragment is affiliated.
- **Geometric Data:** Contains comprehensive digital geometric information of the scanned fresco, including center of mass, dimensions, bounding box limits, position, scale, point cloud vertices and faces, diagonal size, and ground truth transformation concerning its assembled position within the group of pieces.
- **ID:** A unique identification label assigned to each fragment.
- **Link:** Provides an online hyperlink to access and download or parse the digital model of the fragment, along with associated files.
- **RGB File(s):** Lists the paths to high-resolution images utilized for photogrammetry and texture mapping procedures.
- **Raw 3D File(s):** Indicates the paths to the raw 3D scanned files corresponding to the bottom and upper parts of the final 3D model of the fresco.
- **Texture:** Specifies the filenames of low- and high-resolution texture maps, typically stored as *.png* files.
- **Version:** Identifies the acquisition session during which the piece was scanned.
- **Weight:** Denotes the weight of the fragment in grams.

(\* = Fields marked with an asterisk, namely artistic style and fresco family, encompass information collected in collaboration with archaeologists.)

## G  Rendering the associated 2D image of each 3D fragment

As fragments of a fresco, the painted surfaces have unique archaeological value. To enable image-based applications such as puzzle solving, inpainting, filling missing regions, and more, we create a high-quality 2D rendering of each fragment, from which it is possible to map the 3D fragments and associated 3D puzzles to corresponding 2D datasets and puzzles. The rendering pipeline follows two steps:

1. Segment the 3D fragments to isolate the painted surface (see Section G.1) and align the surface normal to one axis;

2. Import the fragment on a prepared 3D scene facing a virtual camera and render an image (see Figure 13).

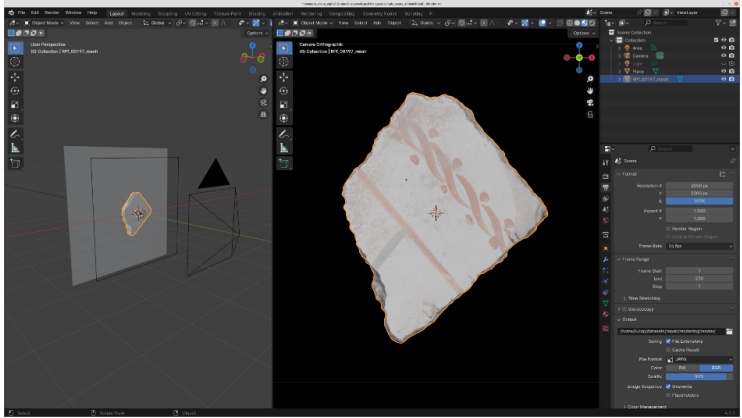

Figure 13: The scene prepared for the rendering of the painted surface of the 3D fragments. On the left, we see the setup with a plane, the fragment and a virtual camera. On the right we see the preview of the 2D image.

The rendering was implemented using a custom scene in Blender using orthographic projection to avoid distortion in the image. To calculate the scaling factor needed to adjust the 3D translation to the 2D translation, we use the orthographic scale and the size of the images. In our set up, we scaled the 3D fragments by $s_{3D} = 0.01$ for easier usage, we rendered them as a square image with $1 : 1$ aspect ratio, $s_{img} = 2000$ pixel resolution, and we orthographic scale of $s_{ort} = 2.714$. We calculate the factor from 3D to 2D as:

$$f_{3D \to 2D} = s_{img} \times s_{3D}/s_{ort} = 2000 \times 0.01/2.714 = 7.369 \ .$$

This allows us to translate the millimeter 3D translation into the 2D pixel translation:

$$T_{2D,px} = T_{3D,mm} * f_{3D \to 2D}$$

Using the 2D pixel translation we can assemble the 2D rendered version of each piece and build the 2D assembled puzzle.

### G.1 Segmenting the 3D fragments and isolating the textured surface

The painted surface of each fragment, unlike its other surfaces, tends to be highly flat. This property, like the sharp corners between most surfaces, can help identify the boundary of the painted surface, as indeed we leverage for its segmentation.

Following the fundamentals of differential geometry, it is understood that each point on a regular surface bends in infinite directions with varying magnitudes. Principal curvatures $\kappa_1$ and $\kappa_2$, along with their perpendicular principal directions, describe the maximum and minimum curvature at each vertex [21] and serve as a descriptor for the local shape. We thus estimate these principal curvatures for all vertices in the surface mesh [60], followed by a computation of their curvedness [40, 47], $\sqrt{(\kappa_1{}^2 + \kappa_2{}^2)/2}$. This measure informs how sharp or smooth the mesh is at each vertex. Having the curvedness map on the mesh, we apply a region-growing algorithm to cluster the fragment's vertices and to sufficiently large surfaces. The smoothest surface (*i.e.*, having the lowest mean of curvedness) is identified as the textured surface.

## H  Ground truth generation

The solution of each puzzle has been manually created following the reference image taken during the digitization phase from the physical assembled solution from the archaeologists.

### H.1  Tool for annotation

We have chosen to use Blender as a tool to annotate the position and orientation of each piece. The choice is motivated by the following reasons:

- any user can use Blender without a strong technical knowledge, enabling people from different fields to contribute to the manual annotation, where archaeological expertise may be more important than computer vision background;

- during annotation, the user sees the assembly from different perspective at the same time, allowing the user to have a better overview and accurately align the pieces (see Figure 14);

- Blender is an open source tool with a strong community, providing high-quality support on long terms, guaranteeing a long-term support for modifying, improving and extending annotations of this dataset without explicit support from the research group after the end of the project;

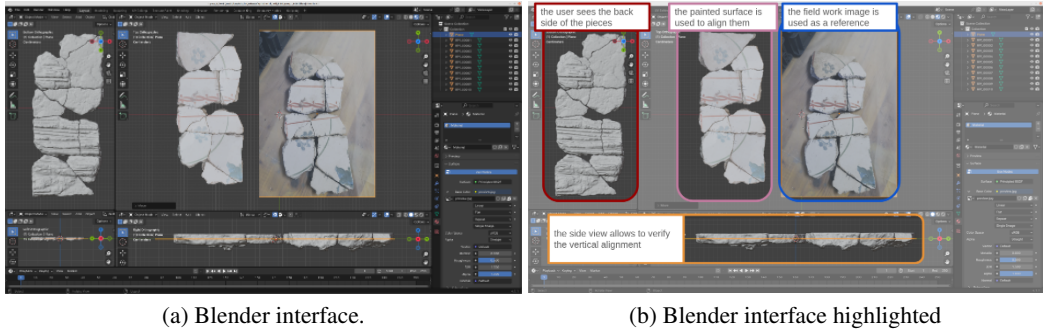

(a) Blender interface.                    (b) Blender interface highlighted

Figure 14: The interface on Blender with the different views.

### H.2  Solving 3D Puzzles

Each puzzle consists in a set of pieces with at least one connection to another piece. Puzzles are typically part of large, flat areas, as they come from parts of ceiling or walls. The assembled solution often show missing fragments, eroded areas and different levels of degradation of the painted surface, as visible in Figure 15. However, due to degradation, the geometry of the pieces may not be perfectly aligned with the neighbouring pieces, and the texture information is crucial for determining and choosing the correct assembly transformation.

As previously mentioned, the annotation is made following a reference image and using the information of the different views to achieve the most accurate placement.

In Figure 15 we show a couple of examples of such puzzles assembled.

## I  Experiment details

### I.1  Evaluation metric

**Definition of the neighbors.** In the 3D implementation, we can directly use the distances between the point cloud of the fragments. While, in the 2D implementation, we chose to dilate all fragments by more than half of the maximal number of pixels or volume (3D) between neighbors in the group; and pair together every two intersecting pieces. This naive approach might fail if a thinner fragment than the dilation radius, separates the two other fragments, but such anomalies were not identified in the dataset.

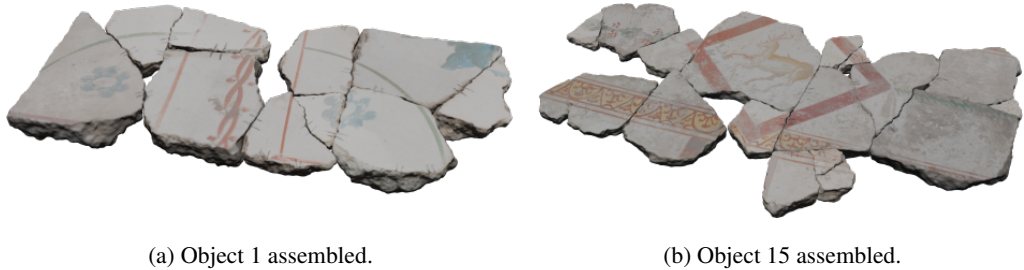

| (a) Object 1 assembled. | (b) Object 15 assembled. |

Figure 15: Some examples of assembled 3D objects. These have been done by a human expert using as a reference the image of the physical reconstruction from archaeologists.

### I.2   2D puzzle

**Hardware.** The experiments were conducted on machines with the following specifications:

- 12th Gen Intel(R) Core(TM) i7-12700K Processor with a base clock speed of 3.60 GHz and 32.0 GB of RAM.
- 12th Gen Intel(R) Core(TM) i7-12700H Processor with a base clock speed of 2.30 GHz and 16.0 GB of RAM.

**Baseline Methods.** We benchmark the performance of a SoTA approach for archaeological puzzles' reassembly; a reconstruction approach based on evolutionary optimization, and a pure geometric reconstruction approach.

- **Derech *et al*. [20] Archaeological Puzzle Solver**. A greedy reconstruction algorithm that iteratively picks the next best additional fragment to add to a growing reconstruction. The scoring procedure is based on calculating dissimilarity between extrapolated parts of potentially neighboring fragments. It should be noted that Derech *et al*. [20] use an outdated extrapolation process that we replaced with the stable-diffusion extrapolation process presented in Harel *et al*. [34].

- **Genetic reconstruction**. This approach employs genetic programming, an optimization technique inspired by the core principle of natural selection – survival of the fittest. Initially, a set of $M$ random reconstructions is generated. A solution for a group with $N$ fragments is represented by an $N \times 3$ matrix, indicating $x$-$y$ coordinates and rotation of each piece. In every generation, the $M/4$ least fit matrices are replaced by the offsprings of the $M/2$ fittest solutions. An offspring is computed by taking the element-wise average of two matrices, followed by a small random mutation of its values, two operations at the foundations of biological evolution. To assess the fitness of a solution, we calculate both the area of the puzzle's bounding rectangle and the intersection area of overlapping pieces. Since perfect solutions tend to minimize both of these values, optimizing their weighted sum guides the solution towards desired reconstructions characterized by a compact yet overlap-free configuration of pieces. Naturally, this does not guarantee a good solution, and failures are possible.

- **Greedy geometric matching**. This puzzle solver initializes a reconstruction by choosing randomly a seed fragment and then iteratively extends the plaque in a greedy fashion similar to the Derech *et al*. [19] baseline. However, here the greedy selection is based on geometric compatibility between segments in the fragment contours.
  More specifically, we used the Douglas-Peucker algorithm [23] to compute the polygonal approximation $((x_1, y_1), ..., (x_m, y_m), (x_1, y_1))$ of the fragment contours. For every point $(x_i, y_i)$ in this sequence, we estimate its discrete curvature $\kappa_i$ using standard methods. Polygon vertices whose curvature exceeds a predefined threshold serve to segment the contour to segments, where each segment is thus a sequence of contour points, and each contour is a list of adjacent segments.
  After the segments are computed, the solver follows the spring-mass physical optimization from Harel *et al*. [33, 34] to assess the compatibility of any two segments from two different fragments.

Table 7: Results on RePAIR 3D Puzzle.

| Method | $Q_{pos}$ ↑ | RMSE ($\mathcal{R}$) ↓ | RMSE ($\mathcal{T}$) ↓ | Precision ↑ | Recall ↑ | F1 ↑ |
|---|---|---|---|---|---|---|
| Global [49] | 0.002 | 88.697 | 91.385 | 0.699 | 0.862 | 0.746 |
| LSTM [90] | 0.005 | 87.601 | 90.097 | 0.668 | 0.813 | 0.697 |
| DGL [94] | 0.002 | 78.467 | 90.531 | 0.674 | 0.792 | 0.706 |
| SE(3)-Equiv. [89] | 0.018 | 84.759 | 94.213 | 0.694 | 0.775 | 0.714 |
| DiffAssemble [67] | 0.384 | 66.781 | 90.010 | 0.521 | 0.836 | 0.629 |

This is done by anchoring two springs to the endpoints of the pair of tested segments and minimizing the system's energy while allowing overlap between the fragments. Upon convergence, the compatibility score is the intersection over the union of the two bodies.

### I.3 3D puzzle

**Hardware.** The experiments were conducted on one machine: an Intel(R) Xeon(R) Silver 4210 CPU @ 2.20GHz Sky Lake with 380 GB RAM, equipped with four NVIDIA Tesla V100 16GB.

**Model Settings.** We train all the models for 500 epochs, using the default learning rate and optimization algorithm for each baseline.

**Baseline Methods.** These are additional details on the baselines we used for the experimental evaluation of the 3D tasks.

- **Global [49]**. Features are extracted for each part from the input point cloud, along with a global feature, using PointNet [65]. Subsequently, the global feature is concatenated with the individual part features and passed through an MLP network with shared weights to estimate the SE(3) pose for each input point cloud.

- **LSTM [90]**. A bi-directional LSTM [69] module is created to enhance the understanding of relationships between parts, with part features being processed and the SE(3) pose for each input point cloud sequentially predicted. This approach mirrors the step-by-step decision-making process utilized by humans during shape assembly.

- **DGL [94]**. GNNs capture part features through modules that reason over edge relationships and aggregate information from nodes. The node aggregation step is removed, which was originally designed to handle geometrically equivalent parts in DGL. This decision is made due to the distinct geometric properties of each piece in the dataset.

- **SE(3)-Equiv. [89]**. Takes a point cloud for each part and generates two representations: equivariant and invariant. It then computes correlations between parts to create an equivariant representation for each part. Using these representations, rotation and translation decoders predict each part's pose. Performance is further enhanced through additional techniques like adversarial training and reconstructing a canonical point cloud.

- **DiffAssemble [67]**. The pieces are modeled as nodes in a complete graph, with each node having its own feature encoder. Time-dependent noise is introduced to the translation and rotation of each piece to simulate a shuffled state, similar to scattering puzzle pieces or 3D fragments. During training, an Attention-based GNN processes this noisy graph to restore the original translation and rotation of the pieces. During inference, The pieces are randomized in positions and rotations, and the noise is removed iteratively to reassemble the pieces.

### I.4 Additional ablation study on 3D puzzle solving

By default, differently from the 2D baselines, the 3D methods do not include an anchored fracture. For this reason, we also explore this solution for the 3D baselines: we establish the coordinate system based on the largest piece and ensure that this piece remains fixed without any movement in translation or rotation.

Table 7 reports the results of the baseline with the pivot. Comparing with Table 4, it reveals a significant degradation in performance across all baselines for translation. However, in rotation, three out of five baselines (DGL, SE(3)-Equiv to DiffAssemble) exhibit improvement. Regarding $Q_{pos}$, performance worsens except for DiffAssemble, while all baselines show enhancement in F1 scores.

Hence, these findings suggest that employing the pivot enhances fracture rotation despite a deviation from the ground truth position, the latest outcome can also be confirmed in Figure 18 Also for this experimental setting, our results indicate a significant gap in resolving the problem.

Additional qualitative results based on the initial configuration of the 3D evaluted baselines can be depicted in Figures 16 and 17 respectively.

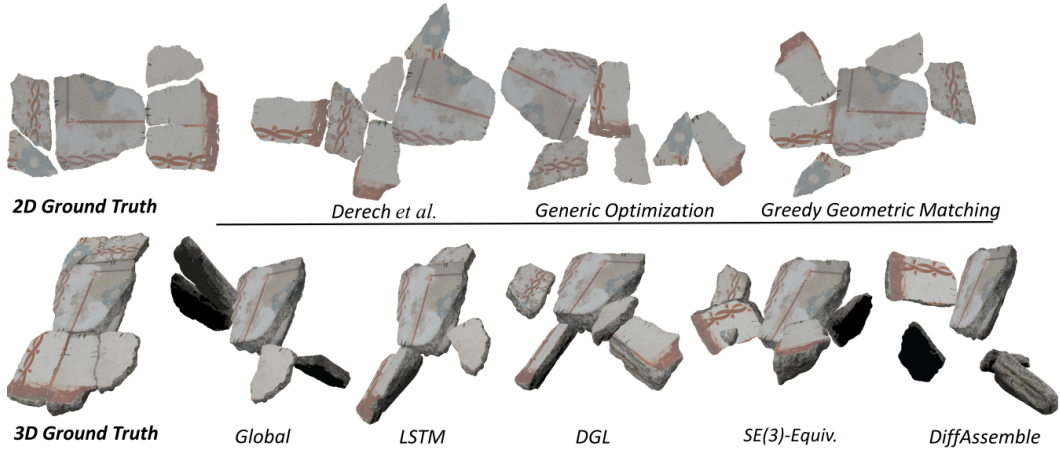

Figure 16: 2D and 3D qualitative results in one of the RePAIR dataset groups.

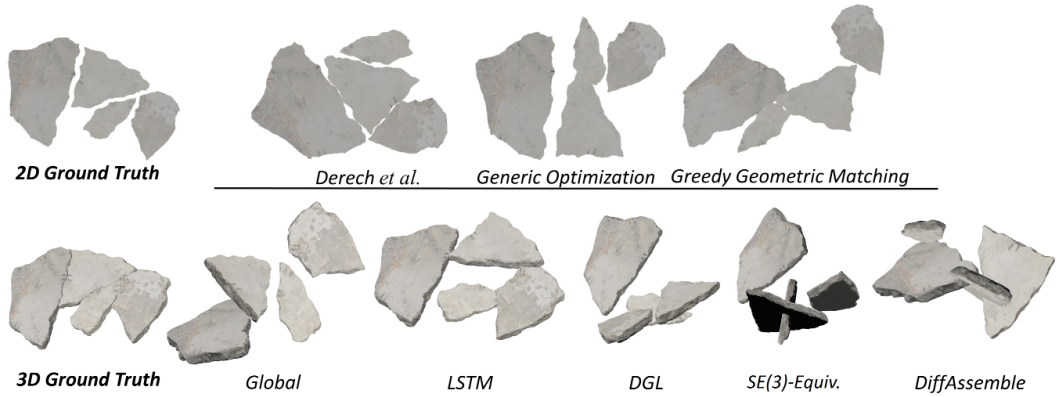

Figure 17: 2D and 3D qualitative results in one of the RePAIR dataset groups.

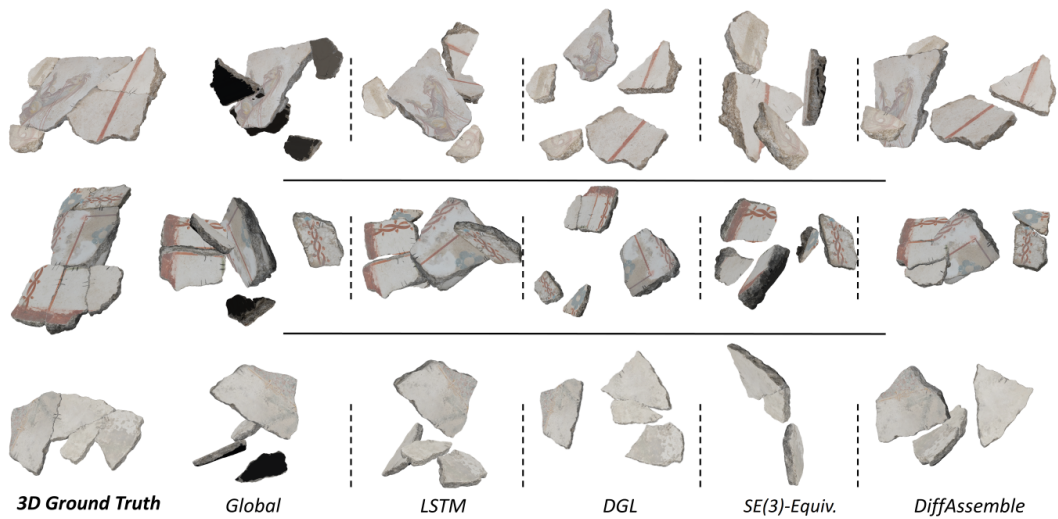

**3D Ground Truth**  *Global*  *LSTM*  *DGL*  *SE(3)-Equiv.*  *DiffAssemble*

Figure 18: Additional 3D qualitative results in the same groups depicted in the previous figures but including an anchored fracture as a reference piece.